\documentclass[10pt,twocolumn,letterpaper]{article}

\usepackage[pagenumbers]{cvpr} 

\usepackage{graphicx}
\usepackage{amsmath}
\usepackage[accsupp]{axessibility}
\usepackage{amssymb}
\usepackage{booktabs}
\usepackage[noend]{algpseudocode}

\usepackage{algorithmicx,algorithm}

\usepackage{times}
\usepackage{epsfig}
\usepackage{pifont}       
\usepackage{bbding}       
\usepackage{setspace}
\usepackage{multirow}
\usepackage{cite}
\usepackage{comment}
\usepackage{booktabs}
\usepackage{arydshln}
\usepackage{color}
\usepackage{xcolor}
\usepackage{bm}
\usepackage[normalem]{ulem}

\usepackage[pagebackref,breaklinks,colorlinks]{hyperref}

\usepackage[capitalize]{cleveref}
\crefname{section}{Sec.}{Secs.}
\Crefname{section}{Section}{Sections}
\Crefname{table}{Table}{Tables}
\crefname{table}{Tab.}{Tabs.}

\begin{document}

\title{Continual Novel Class Discovery via Feature Enhancement and
Adaptation}

\author{
Yifan Yu,
Shaokun Wang,
Yuhang He{\thanks{\* Corresponding authors}},
Junzhe Chen,
Yihong Gong\\ 
Xi'an Jiaotong University\\
{\tt\small \{yyf1999,cjz2183612038\}@stu.xjtu.edu.cn} \\
{\tt\small shaokunwang.xjtu@gmail.com}, {\tt\small heyuhang@xjtu.edu.cn}, {\tt\small ygong@mail.xjtu.edu.cn}}
\maketitle

\begin{abstract}
Continual Novel Class Discovery (CNCD) aims to continually discover novel classes without labels while maintaining the recognition capability for previously learned classes. 
The main challenges faced by CNCD include the feature-discrepancy problem, the inter-session confusion problem, \emph{etc}. 
In this paper,
we propose a novel Feature Enhancement and Adaptation method for the CNCD to tackle the above challenges, which consists of a guide-to-novel framework, a centroid-to-samples similarity constraint (CSS), and a boundary-aware prototype constraint (BAP).
More specifically, 
the guide-to-novel framework is established to continually discover novel classes under the guidance of prior distribution. 
Afterward, the CSS is designed to constrain the relationship between centroid-to-samples similarities of different classes, thereby enhancing the distinctiveness of features among novel classes. 
Finally, 
the BAP is proposed to keep novel class features aware of the positions of other class prototypes during incremental sessions, and better adapt novel class features to the shared feature space. 
Experimental results on three benchmark datasets demonstrate the superiority of our method, especially in more challenging protocols with more incremental sessions.
\end{abstract}

\begin{figure}[t]
\centering
\includegraphics[width=1\linewidth]{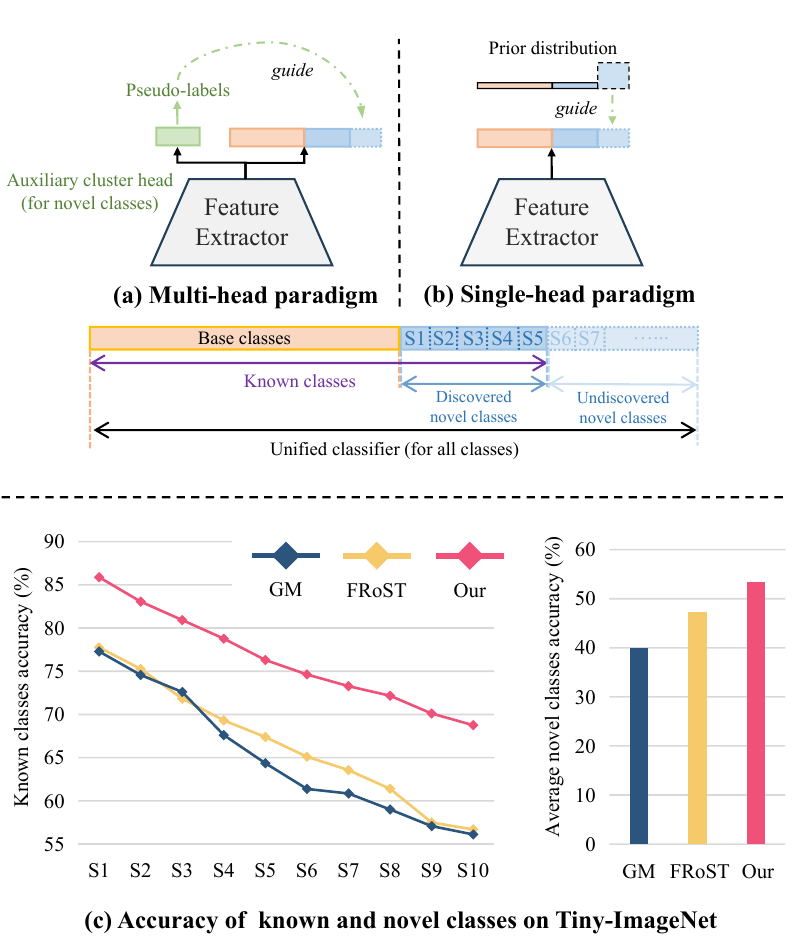}

\caption
{Comparison of the multi-head paradigm (GM, FRoST, \emph{etc}.) and our single-head paradigm to tackle the feature-discrepancy problem. (a) The multi-head paradigm requires pseudo-labels from an auxiliary cluster head as self-supervision signals. (b) Our single-head paradigm uses prior distribution as a self-supervision signal. (c) Accuracy comparison of different methods on Tiny-ImageNet with 10 incremental discovery sessions. The left figure illustrates the accuracy of known classes at the end of each incremental session, and the right figure shows the average accuracy of novel classes across all sessions. ‘S1’ represents the 1-st incremental session.}

  \label{pic:compare}

\end{figure}

\section{Introduction}
Despite the rapid progress of DNNs in computer vision tasks~\cite{carion2020end,wang2023non, Dong_2023_CVPR,lin2023uer,sheng2021improving}, most of them focus on a simple environment where the classes to be identified are pre-defined in a closed set~\cite{boult2019learning}. 
In real-world applications, however, the DNNs are exposed to dynamic environments with a continual emergence of novel classes without any supervision. There is an urgent need to liberate DNNs from the constraint of the closed-set assumption, allowing them to continually discover novel classes without labels while also maintaining the recognition capability for known classes, \emph{i.e.}, the Continual Novel Class Discovery (CNCD) task~\cite{roy2022class,chen2024adaptive}. 
The goals of the CNCD are twofold: effectively discovering novel classes without labels and addressing the catastrophic forgetting problem~\cite{french1999catastrophic} of known classes learned through supervised training.

The CNCD is faced with two major challenges: (1) The feature-discrepancy problem. 
The class-discriminative features obtained through supervised training are suitable for classification while discovering novel classes without labels requires more diverse and rich features~\cite{zhang2022grow}.
Under the premise of maintaining the classification performance of known classes, the discrepancy between two types of features hinders the model's ability to effectively discover novel classes. (2) The inter-session confusion problem~\cite{survey1}.
The features of known and novel classes are trained separately in different training sessions, making it difficult to establish sharp decision boundaries among all classes.  To address the feature-discrepancy problem, existing CNCD methods follow a multi-head paradigm. 
As shown in Figure~\ref{pic:compare}(a), an auxiliary cluster head is exploited exclusively for discovering novel classes in unsupervised ways and assigning pseudo-labels for the unlabeled novel class samples. 
With these pseudo-labels, they train a unified classifier to learn the novel classes.
The pseudo-labels, however, are prone to errors, and their one-hot format provides scanty information, limiting the long-term capability for discovering novel classes, as shown in Figure~\ref{pic:compare}(c). 
Moreover, to mitigate the inter-session confusion, some CNCD methods adopt a rehearsal-based strategy. GM~\cite{zhang2022grow} stores representative samples (\emph{i.e.}, exemplars) for known classes and replays them in the later incremental discovery sessions, facilitating the mutual adaptation of features between novel and known classes. 
Despite their promising performances, storing class exemplars is often infeasible in real-world scenarios due to memory limits and privacy concerns. 
Other CNCD methods adopt a prototype-based strategy. FRoST~\cite{roy2022class} calculates the mean of sample features (\emph{i.e.}, prototype) for each known class and then generates pseudo-features in later sessions to alleviate catastrophic forgetting.
Nonetheless, when discovering novel classes with the auxiliary cluster head, it overlooks the positions of known class prototypes in the shared feature space. The novel class features will be mixed with pseudo-features of known classes, leading to ambiguous decision boundaries between the novel and known classes.

To meet these challenges, we propose a novel Feature Enhancement and Adaptation (FEA) method for the CNCD task. 
The proposed FEA method consists of a guide-to-novel framework, a centroid-to-samples similarity constraint (CSS), and a boundary-aware prototype constraint (BAP). 
First,
to address the feature discrepancy problem, as shown in Figure~\ref{pic:compare}(b), we propose a guide-to-novel framework following a single-head paradigm, primarily composed of a feature extractor and a unified classifier.
Specifically, for discovering novel knowledge, a guide-to-novel term is designed to align the model’s predicted class distribution with the prior distribution. This alignment guides the unified classifier to discover novel classes from rich features obtained through contrastive learning. 
In addition, we adopt a prototype-based strategy to maintain old knowledge. 
Second, to enhance the discriminability of features among novel classes, we design the centroid-to-samples similarity constraint. 
Concretely, we regard the similarity between a novel class centroid and all samples in a mini-batch as a kind of class representation. 
By constraining the relations of all novel class representations, sharp decision boundaries can be established among novel classes. 
Third, 
to alleviate inter-session confusion, we develop the boundary-aware prototype constraint. 
By keeping novel class features aware of the positions of other class prototypes during incremental sessions, this constraint can reduce the interference of novel class features on the decision boundaries of known classes. 
Afterward, the novel class features can be well adapted to the shared feature space.  
We conduct extensive experiments on three benchmark datasets, including CIFAR-10, CIFAR-100, and Tiny-ImageNet, using more challenging and practical experimental protocols. We are the first to apply CNCD to long-incremental protocols (up to ten sessions), whereas previous works use only two or three sessions. The experimental results demonstrate that the FEA significantly outperforms existing CNCD methods, confirming its effectiveness and superiority, especially with more incremental sessions. Our main contributions can be summarized as follows: 

\begin{itemize}
\item For the CNCD task, we propose a novel Feature Enhancement and Adaptation (FEA) method, which consists of a guide-to-novel framework, a CSS constraint, and a BAP constraint. 
Experimental results on three benchmark datasets demonstrate the superiority of our FEA method, especially with more incremental sessions.


\item We establish a guide-to-novel framework following a single-head paradigm. 
This framework can continually discover novel classes from rich features obtained through contrastive learning with the guidance of a robust prior distribution, thereby addressing the feature-discrepancy problem.

\item For feature enhancement, we design a centroid-to-samples similarity constraint to enhance the distinctiveness of features among novel classes, which promotes the effective discovery of novel classes.

\item For feature adaptation, we design a boundary-aware prototype constraint to better adapt novel class features to the shared feature space, thereby alleviating the inter-session confusion problem.

\end{itemize}

\begin{figure*}[t]
	\centering

	\includegraphics[width=1\textwidth]{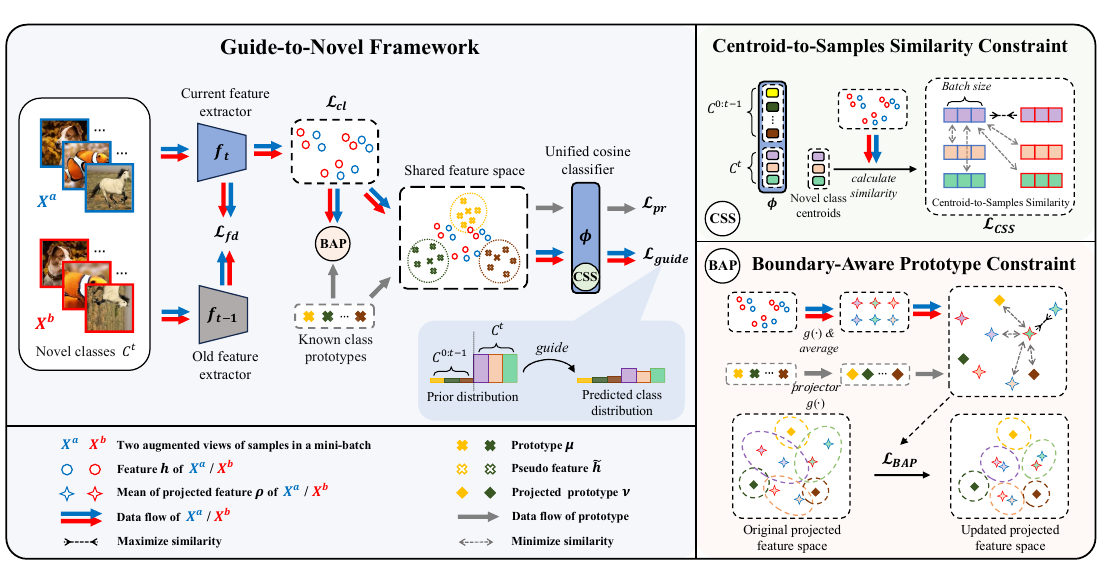} 

	\caption{
	An overview of our FEA for CNCD. The proposed FEA is comprised of a guide-to-novel framework, a centroid-to-samples similarity constraint (CSS), and a boundary-aware prototype constraint (BAP). The guide-to-novel framework can continually discover novel knowledge while maintaining old knowledge without any auxiliary cluster head. The CSS is designed to enhance the distinctiveness between novel class features. The BAP is proposed to better adapt novel class features to the shared feature space.
	} 

	\label{figs:framework}

\end{figure*}

\section{Related Work}
\subsection{Class-Incremental Learning}
Class-Incremental Learning (CIL) aims to sequentially learn new classes while maintaining the performance of previously encountered classes. 
To mitigate the catastrophic forgetting problem~\cite{french1999catastrophic}, existing methods fall into three major categories:
(1) Regularization-based approaches design regularization terms to penalize the model for forgetting old knowledge~\cite{kirkpatrick2017overcoming,zhang2020class,wu2021striking}.
(2) Architecture-based approaches focus on how to modify the network structure to integrate knowledge acquired from both new and old classes~\cite{hung2019compacting,yan2021dynamically,kim2023open}.
(3) Rehearsal-based approaches store representative samples and replay them in the following incremental sessions to alleviate catastrophic forgetting. For instance, iCaRL~\cite{rebuffi2017icarl} uses the herding algorithm~\cite{welling2009herding} for exemplar selection and a nearest-mean-of-exemplars classifier for classification. 
A large number of methods are based on rehearsal-based strategy and design their regularization losses~\cite{castro2018end,hou2019learning}. 
However, due to the limited storage and privacy issues, storing old class exemplars is infeasible in some application scenarios. 
Therefore, some methods employ a prototype-based approach which does not require the storage of exemplars. 
PASS~\cite{zhu2021prototype} and SSRE~\cite{zhu2022self} store prototypes (\emph{i.e.}, the mean of class features) for old classes and then generate pseudo features by using these prototypes to jointly train with new class features in incremental sessions. 
In this paper, we employ the prototype-based strategy to alleviate forgetting for known classes and design a boundary-aware prototype constraint to make novel class features adapt better to the shared feature space.

\subsection{Continual Novel Class Discovery}
The goal of Novel Class Discovery (NCD) is to utilize prior knowledge driven from known classes to cluster novel classes in unlabeled data, which is formalized by DTC~\cite{Han2019learning}. 
In the NCD setting, there is no overlap between known (\emph{i.e.}, labeled) classes and novel (\emph{i.e.}, unlabeled) classes.
Up to now, numerous NCD methods can be divided into two categories: one-stage approaches and two-stage approaches~\cite{troisemaine2023novel}. 
(1) One-stage approaches can leverage both the labeled and unlabeled data simultaneously to train a shared feature space~\cite{fini2021unified,Gu_2023_ICCV,li2023modeling,yang2023bootstrap}. For instance, AutoNovel~\cite{han2021autonovel} adopts self-supervision learning to pre-train the model with all data and proposes a ranking statistics method for pseudo labeling. 
However, this kind of approach is not suitable for incremental scenarios. 
(2) Two-stage approaches are similar to cross-task transfer learning, which first learns on labeled data and then migrates to unlabeled data for novel class discovery, such as L2C~\cite{Hsu18_L2C} and MCL~\cite{Hsu19_MCL}. Although two-stage approaches have an embryonic form of CNCD, they only care about the accuracy of unlabeled novel data, not that of labeled known data. 
Based on this, the Continual Novel Class Discovery (CNCD) emerges, which simultaneously focuses on the accuracies of both novel and known data. 
FRoST~\cite{roy2022class} and GM~\cite{zhang2022grow} train auxiliary cluster heads exclusively for novel classes in unsupervised ways and use them to assign pseudo-labels to guide the training of a unified classifier.
NCDwF~\cite{joseph2022novel} uses pseudo-labels generated by Sinkhorn-Knopp algorithm~\cite{cuturi2013sinkhorn} to train an auxiliary cluster head and designs a binary known class identifier for the task-agnostic inference, which can not handle multiple incremental sessions. 
These methods all follow the multi-head paradigm for discovering novel classes, but pseudo-labels from auxiliary cluster heads are unreliable, and their one-hot format is unable to provide rich information.
In this paper,
to address the above issues, we propose a guide-to-novel term as a robust self-supervision signal and the centroid-to-samples similarity constraint to promote the effective discovery of novel classes.

\section{Method}
\subsection{Problem Definition}
In the setting of CNCD, training sessions can be divided into a base session ($i.e.$, 0-th session) and $T$ incremental discovery sessions. 
At the base session, the model is initially trained on the labeled dataset $\mathcal{D}^{l}=\left \{ \left ( x_{i}^{l}, y_{i}^{l}  \right )  \right \}_{i=1}^{\left | \mathcal{D}^{l} \right |} $, where label $y_{i}^{l}\in \mathcal{C}^{0}$ and $ \mathcal{C}^{0}$ is the class set of $\mathcal{D}^{l}$. 
Afterward, 
the model sequentially discoveries novel classes on a series of unlabeled datasets $\mathcal{D}^{u} =\left \{  \mathcal{D}_{t}^{u}\right \} _{t=1}^{T} $, where $\mathcal{D}_{t}^{u} =\left \{  x_{i,t}^{u}\right \} _{i=1}^{\left | \mathcal{D}_{t}^{u} \right | }$ is the unlabeled data at $t$-th incremental discovery session and the corresponding ground-truth labels are unknown.
Following the NCD settings, class sets from different sessions are disjoint, $i.e.$, $\mathcal{C}^{0}\cap \mathcal{C}^{1}\cap\mathcal{C}^{2}\cap\cdots \cap\mathcal{C}^{t}=\phi $, where $\mathcal{C}^{t}$ is the class set of $\mathcal{D}_{t}^{u}$. 
Note that previous datasets are unavailable at the current incremental training session and all classes to be discovered at the current incremental training session are novel classes. Once a novel class is discovered, we treat it as a known class. The goal of CNCD is to continually discover novel classes $\mathcal{C}^t$ without labels while maintaining the ability to classify known classes $\mathcal{C}^{0:t-1}= \mathcal{C}^{0}\cup \mathcal{C}^{1}\cup\mathcal{C}^{2}\cup\cdots \cup\mathcal{C}^{t-1}$.
\paragraph{\bf Overview of Method.} Our method is shown in Figure~\ref{figs:framework}, including a guide-to-novel framework following the single-head paradigm, a centroid-to-samples similarity constraint (CSS), and a boundary-aware prototype constraint (BAP).

\subsection{A Guide-to-Novel Framework}
Our guide-to-novel framework comprises a feature extractor $f(\cdot)$ and a unified cosine classifier $\phi(\cdot)$. At the base session, we train our model on the labeled dataset $\mathcal{D}^{l}$ by the standard cross-entropy loss $ \mathcal{L}_{ce}$:
\begin{equation}
  \mathcal{L}_{ce}(y_i^l,p_i)=-\sum_{j\in \mathcal{C}^{l}}y_{i,j}^l\log(p_{i,j}),
\end{equation}
where $y_i^l$ is the ground-truth label and the predicted probability $p_i=softmax(\phi(f(x_i^l)))$. 
At the beginning of $t$-th incremental discovery session, the unified cosine classifier $\phi_t(\cdot)$ is first extended to accommodate all known and novel classes.
\subsubsection{\bf Maintaining old knowledge.} 
We adopt the prototype-based strategy to alleviate catastrophic forgetting. 
At the end of the last session $t$-1, we store the feature prototype $\mu_c=\frac{1}{n_c} \sum_{i=1}^{n_c} h_i$ and variance $\sigma_c^2 =\frac{1}{n_c} \sum_{i=1}^{n_c}(h_i-\mu_c)^2$ for each known class, where $n_c$ is the number of samples whose ground-truth (or predicted) labels are class $c$ and $h_i=f(x_i)$ is the feature of $x_i$. 
At session $t$, pseudo features of known class $c$ can be generated by $\tilde{h}_c=\mu_c+e$ ,where $e\sim \mathcal{N} (0,\sigma_c)$ is the Gaussian noise. 
Then, the pseudo rehearsal loss $\mathcal{L}_{pr}$ is defined as:
\begin{equation}
\mathcal{L}_{pr}=\frac{1}{\left | \mathcal{C}^{0:t-1} \right |} \sum_{c\in\mathcal{C}^{0:t-1}}\frac{1}{n_p}\sum_{i=1}^{n_p} \mathcal{L}_{ce}(y_{i,c},\tilde{p} _{i,c}),
\end{equation}
where $n_p$ is the number of generated pseudo features for each known class, $y_{i,c}$ is the label of pseudo feature $\tilde{h}_{i,c}$, and $\tilde{p}_{i,c}=softmax(\phi(\tilde{h}_{i,c}))$ is the predicted probability for $\tilde{h}_{i,c}$. 
Finally, the loss to maintain old knowledge is formulated as follows:
\begin{equation}
\mathcal{L}_{old}=\mathcal{L}_{pr}+\frac{\lambda}{\left |\mathcal{B}^u \right | } \sum_{i\in\mathcal{B}^{u}} \mathcal{L}_{fd}(f_{t-1}(x^u_{i,t}),f_{t}(x^u_{i,t})),
\end{equation}
where $\mathcal{B}^u$ is a mini-batch from $\mathcal{D}^{u}_t$, $\mathcal{L}_{fd}$ is the feature distillation loss ~\cite{KD}, $f_{t-1}(\cdot)$ and $f_{t}(\cdot)$ are feature extractors at session $t$-1 and $t$,  respectively, and $\lambda$ is a weighting hyper-parameter. $\mathcal{L}_{pr}$ and $\mathcal{L}_{fd}$ are employed to mitigate the catastrophic forgetting of our framework. 
\subsubsection{\bf Discovering novel knowledge.}
Due to the insufficient exploration of rich information in images by pseudo-labels and the potential presence of wrong pseudo-labels, we employ contrastive learning~\cite{chen2020simple} and our proposed guide-to-novel term $D_{g}$ to address these issues, respectively.

\paragraph{\bf Enriching feature diversity.}
Features learned through contrastive learning are more suitable for downstream tasks compared to those learned through supervised learning~\cite{tian2020makes,wu2023metagcd}.
The old feature space is class-discriminative for known classes, which hinders the model’s ability to effectively discover
novel classes without labels. Thus, to provide diverse and informative features for CNCD, the self-supervised contrastive loss $\mathcal{L}_{cl}$ is adopted to enrich features for novel classes. Concretely, for each unlabeled sample $x_{i,t}^{u}$ in a mini-batch $B^u$, random augmentation is performed twice to get two augmented views (\emph{i.e.}, $ x_{i,t}^{a}$ and $ x_{i,t}^{b}$) of the sample. 
The $\mathcal{L}_{cl}$ is formulated as:
\begin{equation} 
    \resizebox{0.99\linewidth}{!}{$
            \displaystyle
\mathcal{L}_{cl} =-\frac{1}{\left |\mathcal{B}^u \right |}\sum_{i\in\mathcal{B}^{u} }
\log\frac{\exp(\bar{z}_{i}^a \cdot \bar{z}_{i}^b)}
{ {\textstyle \sum_{j\in\mathcal{B}^{u} }}\left [  \exp(\bar{z}_{i}^a \cdot \bar{z}_{j}^a)+\exp(\bar{z}_{i}^a \cdot \bar{z}_{j}^b)\right ] }
        $},
\end{equation}
where $\mathcal{B}^u$ is a mini-batch from $\mathcal{D}^{u}_t$, the projected feature $z_{i}^a={MLP}(f(x_{i,t}^a))$, and $\bar{z}_{i}^a=z_{i}^a/\left \|z_{i}^a\right \|_2$.
\paragraph{\bf Guiding novel class discovering.}
To guide a unified cosine classifier in discovering novel classes from diverse and informative features, we design a guide-to-novel term $D_{g}$ based on the KL divergence:
\begin{equation} 
D_{g}=D_{KL}(p^r\left | \right |\bar{p}),
\end{equation}
with
\begin{equation}
p_i^r=\begin{cases}
0, & 0<i\le\left | \mathcal{C}^{0:t-1} \right |; \\ 
\frac{1}{\left | \mathcal{C}^{t} \right |}, &\left | \mathcal{C}^{0:t-1} \right |<i\le \left | \mathcal{C}^{0:t} \right |,
\end{cases}
\end{equation}
where $p^r$ is a prior distribution of current novel classes $\mathcal{C}^t$, indicating that samples of the current session $t$ only belong to $\mathcal{C}^t$, and $\bar{p}=\frac{1}{2 \left | \mathcal{B}^{u} \right |} {\textstyle \sum_{i\in\mathcal{B}^{u} }}(p^a_i+p^b_i)$ is the predicted class distribution of mini-batch samples, where $p^a_i$ and $p^b_i$ are the predicted probabilities corresponding to $ x_{i,t}^{a}$ and $ x_{i,t}^{b}$. 
Unlike the inaccurate self-supervision signals provided by pseudo-labels, $D_{g}$ provides a more robust distribution-level self-supervision signal.
More specifically, $D_{g}$ can guide the classifier to acquire knowledge about novel classes by aligning the model's predicted class distribution with the prior distribution of novel classes. 
Formally, we train the unified cosine classifier with a cross-entropy loss $ \mathcal{L}_{ce}$ and the guide-to-novel term $D_{g}$:
\begin{equation}
  \mathcal{L}_{guide} =\frac{1}{\left |\mathcal{B}^u\right |}\sum_{i\in\mathcal{B}^{u} }\mathcal{L}_{ce}(p^a_i ,p^b_i )+ \varepsilon D_{g},
\end{equation}
where $\varepsilon$ is a weighting hyper-parameter. Then, the loss to discover novel knowledge is formulated as:
\begin{equation}
\mathcal{L}_{novel}=\mathcal{L}_{cl}+\mathcal{L}_{guide},
\end{equation}
where $\mathcal{L}_{cl}$ and $\mathcal{L}_{guide}$ are used to enable the feature extractor and the cosine classifier to learn novel classes, respectively.

Finally, the loss function for our guide-to-novel framework is as follows:
\begin{equation}
\mathcal{L}_{framework}=\mathcal{L}_{old}+\mathcal{L}_{novel}.
\end{equation}

In this way, our single-head paradigm framework addresses the feature discrepancy problem, that is, effectively discovering novel classes without labels under the premise that the known class features maintain class-discriminative.

\begin{table*}
\caption{Comparison of the average accuracy with the SOTA methods on CIFAR-10, CIFAR-100 and Tiny-ImageNet. SimGCD$^\ast$ denotes the original SimGCD enhanced with $D_{g}$ and iCaRL. FRoST$^\dagger$ denotes the value calculated by using the accuracy of the known and novel classes reported in the original FRoST paper, whose backbone is ResNet-18.}
\renewcommand{\arraystretch}{1.25}
\centering
\resizebox{1.9\columnwidth}{!}{
\begin{tabular}{c|c|cc|cc|ccc|c} 
\hline
\multirow{2}{*}{\textbf{Metric}} 
& \multirow{2}{*}{\textbf{Method}}
& \multicolumn{2}{c|}{\textbf{CIFAR-10}} 
& \multicolumn{2}{c|}{\textbf{CIFAR-100}} 
& \multicolumn{3}{c|}{\textbf{Tiny-ImageNet}} 
& \multirow{2}{*}{\textbf{\# Exemplar}}  \\ 
\cline{3-9}
&             & T=1 & T=2     & T=2   & T=5   & T=2 & T=5 & T=10    &                                       \\ 
\hline
\multirow{6}{*}{\begin{tabular}[c]{@{}c@{}}$Average$\\$Accuracy \uparrow$\end{tabular}}
& GM                 & 85.42  &   88.05   & 76.27 & 69.47 & 73.66  & 68.31 &    66.86     & \multirow{2}{*}{20 samples per class}    \\
& SimGCD$^\ast$      & 83.80  &   80.21  & 81.06 & 74.75 & 75.53  & 70.85 &    70.40    &                                          \\ 
\cline{2-10}
& AutoNovel          & 64.93 & 54.17 & 64.27 & 46.24 & 64.86  & 41.65 &    34.50     & \multirow{4}{*}{0 sample per class}      \\
& FRoST$^\dagger$    &78.39  &   -     & 55.14 &    -  & 51.70  &    -  &   -     &                                         \\
& FRoST              & 78.81&  86.86  & 81.17 & 73.26 & 74.31  & 69.56 &   67.93      &                                         \\ 
\cline{2-9}

& Our                & \textbf{92.66} & \textbf{92.30}  & \textbf{83.44} & \textbf{79.09} &  \textbf{78.48} & \textbf{74.67}  &   \textbf{75.85}     &                                          \\
\hline
\end{tabular}
}

\label{tab:acc}
\end{table*}

\subsection{Feature Enhancement: Centroid-to-Samples Similarity Constraint}
In our guide-to-novel framework, the contrastive loss $\mathcal{L}_{cl}$ and the guide-to-novel term $D_{g}$ provide instance-level and distribution-level self-supervision signals, respectively. 
Nevertheless,
our goal is not only to discover novel classes but also to enhance their distinctiveness from each other, striving for sharp decision boundaries among novel classes.
Therefore, for feature enhancement,
we propose a centroid-to-samples similarity constraint (CSS) as the class-level self-supervision signal. 
The cosine classifier essentially calculates the cosine similarity between the sample features and the centroids (\emph{i.e.}, the classifier weights) of each class cluster. 
We regard the similarity between the class centroid and all samples within the same mini-batch as a kind of class representation.

The similarity between the centroid of novel class $k$ ($k \in \mathcal{C}^{t}$) and augmented views of samples generated from augmentation $a$ in a mini-batch $\mathcal{B}^u$ is defined as follows:
\begin{equation}
s^{a}_k=\left [ s^{a}_{k,1},s^{a}_{k,2},...,s^{a}_{k,i},...,s^{a}_{k,\left |\mathcal{B}^u  \right | } \right ]^\top,
\end{equation}
where $i\in\left [1,\left |\mathcal{B}^u  \right |\right ] $ and $s^{a}_{k,i}=\left (  w_k/\left \| w _k \right \|_2   \right )\cdot \left ( h_i^a/\left \| h_i^a \right \|_2  \right )$ denotes the cosine similarity between class $k$ centroid $w_k$ and the $i$-th sample feature $h^a_i=f(x^a_{i,t})$.
The similarities (\emph{i.e.}, $s_{k}^a$ and $s_{k}^b$) between a single class centroid and two augmented views of mini-batch samples should be close, while the similarities (\emph{i.e.}, $s_{k}^a$ and $s_{j}^a$) between different class centroids and mini-batch samples should be distinct. 
Hence, the centroid-to-samples similarity constraint (CSS) is formulated as:
\begin{equation}
    \resizebox{.99\linewidth}{!}{$
            \displaystyle
\mathcal{L}_{CSS} =-\frac{1}{\left |\mathcal{C}^t\right |}\sum_{k\in \mathcal{C}^{t} }
\log\frac{\exp(\bar{s}_{k}^a \cdot \bar{s}_{k}^b)}
{ {\textstyle \sum_{j\in \mathcal{C}^{t}}}\left [\exp(\bar{s}_{k}^a \cdot \bar{s}_{j}^a)+\exp(\bar{s}_{k}^a \cdot \bar{s}_{j}^b)\right ] }
        $},
\end{equation}
where $\mathcal{C}^t$ is the current class set, and $\bar{s}_{k}^a=s_{k}^a/\left \|s_{k}^a\right \|_2$.

\subsection{Feature Adaptation: Boundary-Aware Prototype Constraint}

To alleviate inter-session confusion, we need to establish sharp decision boundaries in the shared feature space for all classes. It is essential not only to make novel class features more separable but also to ensure that these novel class features better adapt to the shared feature space without conflicting with the features of known classes. 

Toward this end, we design a boundary-aware prototype constraint (BAP):
\begin{equation}
\mathcal{L}_{BAP} =-\frac{1}{\left |\mathcal{C}^t \right |}\sum_{k\in\mathcal{C}^t }
\log\frac{\exp(\bar{\rho}  _{k}^a \cdot\bar{\rho}  _{k}^b)}
{  {\textstyle \sum_{c\in\mathcal{C}^{0:t-1}}} \exp(\bar{\rho}  _{k}^a \cdot \bar{\nu}_c )+\eta }
,
\label{e12}
\end{equation}
with
\begin{equation}
\eta ={{\textstyle \sum_{j\in\mathcal{C}^t }}\left [  \exp(\bar{\rho}  _{k}^a \cdot \bar{\rho}  _{j}^a)+\exp(\bar{\rho}  _{k}^a \cdot \bar{\rho} _{j}^b)\right ]},
\label{m}
\end{equation}
where $\rho_{k}^a$ is the projected prototype for novel class $k$ (calculated as the mean of projected features $g(f(x_{i,t}^a))$ whose predicted label is class $k$),  $\bar{\rho}_{k}^a=\rho_{k}^a/\left \|\rho_{k}^a\right \|_2$, $\nu_c=g(\mu_c)$ is the projected prototype for known class $c$, $\bar\nu_c=\nu_c/\left \| \nu_c\right \|_2$, and $g(\cdot)$ is an MLP projector, which maps known class prototypes and novel class features into a latent space to avoid changing original feature space dramatically.
The first term of the denominator in Eq.(\ref{e12}) is designed to keep the prototypes of novel classes away from the prototypes of known classes. Meanwhile, Eq.(\ref{m}) is utilized to enforce separation among prototypes of different novel classes. 
By employing the boundary-aware prototype constraint (BAP), novel class features gradually converge while being aware of the positions of features of other classes to better adapt to the shared feature space.

\subsection{Optimization Objective}
Finally, the total loss $\mathcal{L}_{FEA}$ of our FEA is formulated as:
\begin{equation}
\mathcal{L}_{FEA}=\mathcal{L}_{framework}+\mathcal{L}_{CSS}+\alpha(\tau)\mathcal{L}_{BAP},
\end{equation}
where $\alpha(\tau)= \omega\cdot\min (\frac{\tau}{e},1)$ is the warm-up function employed to stabilize the training process, $\omega$ represents a weighting hyper-parameter, $\tau$ represents the current epoch, and $e$ represents an inflection point.

\section{Experiments}
\subsection{Experimental Setup}

\paragraph{\bf Datasets.}
We evaluate our method on three common CNCD benchmark datasets: CIFAR-10~\cite{krizhevsky2009learning}, CIFAR-100~\cite{krizhevsky2009learning}, and Tiny-ImageNet~\cite{le2015tiny}. 
CIFAR-10 is composed of 60,000 image samples, divided into 10 distinct classes. Each class contains 5,000 samples for training and 1,000 samples for testing. CIFAR-100 contains 60,000 image samples from 100 classes, with 500 training and 100 testing samples for each class.
Tiny-ImageNet consists of around 120,000 image samples for 200
classes, where each class has 500 training samples, 50 validation
samples, and 50 testing samples. 

\begin{figure*}[t]
	\centering

	\includegraphics[width=1\textwidth]{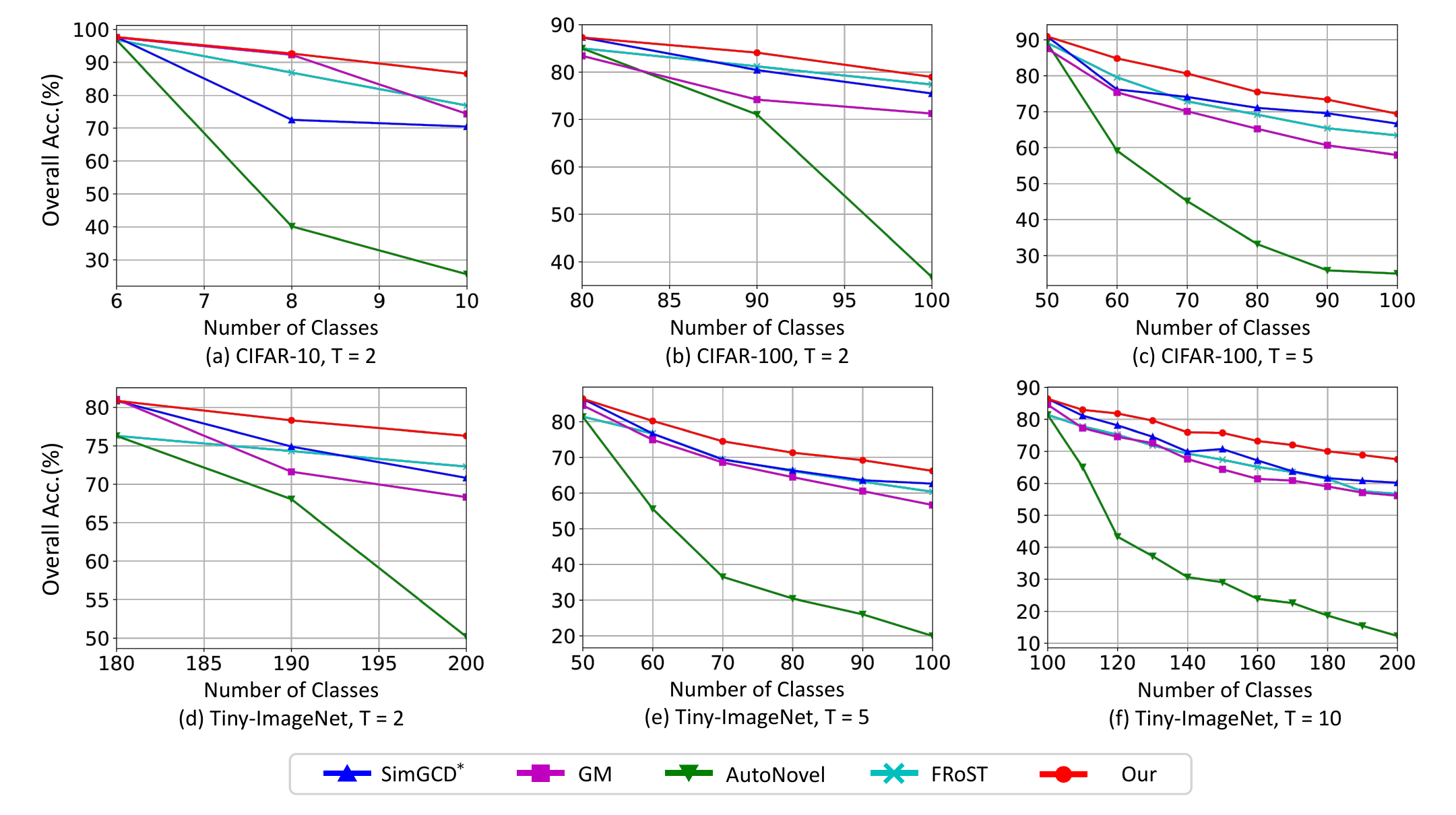} 
	\caption{
The overall accuracy of all learned classes (\emph{i.e.}, known classes) at the end of each incremental discovery session on CIFAR-10, CIFAR-100, and Tiny-ImageNet, where T indicates the number of incremental discovery sessions.}
	\label{figs:acc}
	
\end{figure*}

\begin{table}[]
\caption{Experimental Protocols. We list the total number of classes (\# Class) in the dataset, the number of incremental sessions ($T$), the number of known classes at the base session ($\left |\mathcal{C}^l\right |$), and the number of novel classes at each incremental session ($\left |\mathcal{C}^t\right |$).}
\renewcommand{\arraystretch}{1.25}
\centering
\resizebox{0.80\columnwidth}{!}{
\begin{tabular}{@{}ccccc@{}}
\hline
\textbf{Dataset} & \textbf{\# Class}  & $T$  & $\left |\mathcal{C}^l\right |$  &  $\left |\mathcal{C}^t\right |$  \\ 
\midrule
\multirow{2}{*}{CIFAR-10}     & \multirow{2}{*}{10}  & 1  & 5   & 5  \\ \cmidrule(l){3-5} 
                              &                      & 2  & 6   & 2  \\ \midrule
\multirow{2}{*}{CIFAR-100}    & \multirow{2}{*}{100} & 2  & 80  & 10 \\ \cmidrule(l){3-5} 
                              &                      & 5  & 50  & 10 \\ \midrule
\multirow{3}{*}{Tiny-ImageNet} & \multirow{3}{*}{200} & 2  & 180 & 10 \\ \cmidrule(l){3-5}
                              &                      & 5  & 100 & 20 \\ \cmidrule(l){3-5} 
                              &                      & 10 & 100 & 10 \\ 

\hline
\end{tabular}
}
\label{tab:dataset}

\end{table}

\paragraph{\bf Experimental Protocols.}
In previous works~\cite{roy2022class,zhang2022grow}, the number of incremental sessions is limited (only one or two), which cannot meet the needs of real application scenarios. 
Toward this end, we extend the existing experimental protocols of CNCD with more incremental sessions. Increasing the number of incremental sessions exacerbates the problem of feature discrepancy and inter-session confusion. Thus, long-incremental protocols are more difficult, but have a greater practical application value.

The details of experimental protocols are listed in Table~\ref{tab:dataset}. 
For instance, $T$=5 for Tiny-ImageNet (200 classes) indicates that the first 100 ($\left |\mathcal{C}^l\right |$) classes are used for supervised training at the base session. The remaining 100 classes are divided into 5 ($T$) incremental sessions for unsupervised discovery. And each incremental session has 20 ($\left |\mathcal{C}^t\right |$) novel classes.

\paragraph{\bf Evaluation metrics.}
After all incremental discovery sessions are completed,
we report \textbf{average accuracy}~\cite{rebuffi2017icarl}, \textbf{average forgetting rate}~\cite{Chaudhry_2018_ECCV}, and \textbf{average discovery accuracy} to comprehensively evaluate our method.
In detail, average accuracy represents the mean of overall accuracies across all sessions.
Let $a_{t,j}$ be the class prediction accuracy of the model for classes $\mathcal{C}^j$ belonging to $j$-th session after the training of $t$-th session ($j\le t$). 
The average forgetting rate is written as:
\begin{equation}
F_{t}=\frac{1}{t} {\textstyle \sum_{j=0}^{t-1}} (\max_{l\in \{0,\cdots,t-1\}} a_{l,j}-a_{t,j}), 
\quad \forall j<t.
\end{equation}
Also, the average discovery accuracy is used to evaluate the ability to discover novel classes and is defined as:
\begin{equation}
D_{t}=\frac{1}{t} {\textstyle \sum_{j=1}^{t}}a_{t,j}.
\end{equation}
It is worth mentioning that for unlabeled data from novel classes, cluster accuracy is the maximum accuracy obtained through Hungarian Assignment (HA)~\cite{kuhn1955hungarian}, so it is inappropriate for directly measuring the forgetting of unlabeled data. 
Therefore, after discovering novel classes in the current session, we treat them as known classes and calculate their standard accuracy in subsequent sessions.
Specifically, at the end of each incremental session $t$, we utilize the HA only for current novel classes $\mathcal{C}^t$ to find the optimal permutation from cluster predictions to ground-truth labels, which will be fixed in the following sessions. 
Afterward, with this optimal and fixed permutation, cluster predictions can be transformed into re-assigned predictions, which can be directly compared with ground-truth labels to calculate accuracy. 

\begin{table}
\caption{Comparison of the average forgetting rate and discovery accuracy at the end of the last incremental discovery session with the SOTA methods on CIFAR-10 and CIFAR-100.}
\renewcommand{\arraystretch}{1.3}
\centering
\resizebox{0.9\columnwidth}{!}{
\begin{tabular}{c|c|c|cc} 
\hline
\multirow{2}{*}{\textbf{Metric}} 
& \multirow{2}{*}{\textbf{Method}}
& \textbf{CIFAR-10}
& \multicolumn{2}{c}{\textbf{CIFAR-100}}   \\ 
\cline{3-5}
&             & T=2  & T=2   & T=5               \\ 
\hline
\multirow{6}{*}{\begin{tabular}[c]{@{}c@{}}$AvF. \downarrow$\end{tabular}}
& GM          &  15.43   &  4.28   &  8.09   \\
& SimGCD$^\ast$      &  11.24   &  8.81   &  8.74   \\ 
\cline{2-5}
& AutoNovel   &  90.47   &  41.84   & 49.76  \\
& FRoST$^\dagger$    &-  &  23.11 & - \\
& FRoST       &  11.09   &  3.66  &    7.86      \\ 
\cline{2-5}
& Our         &\textbf{9.98}    &  \textbf{3.12}  &  \textbf{6.92}       \\
\cline{1-5}
\multirow{6}{*}{\begin{tabular}[c]{@{}c@{}}$AvD. \uparrow$\end{tabular}}
& GM          &  62.58   &  54.50       &  41.88   \\
& SimGCD$^\ast$      &  62.68   &   71.90      &  60.10   \\ 
\cline{2-5}
& AutoNovel   &  44.20   &   15.10      &     17.08     \\
& FRoST$^\dagger$    & - &  61.70 &-    \\
& FRoST       &  82.95   &   64.25      &   53.10       \\ 
\cline{2-5}
& Our         &  \textbf{95.63}   &    \textbf{82.65}     &  \textbf{70.86}       \\
\hline
\end{tabular}
}

\label{tab:for_cifar}
\end{table}

\begin{table}
\caption{Comparison of the average forgetting rate and discovery accuracy at the end of the last incremental discovery session with the SOTA methods on Tiny-ImageNet.}
\renewcommand{\arraystretch}{1.25}
\centering
\resizebox{0.9\columnwidth}{!}{
\begin{tabular}{c|c|ccc} 
\hline

\multirow{2}{*}{\textbf{Metric}} 
& \multirow{2}{*}{\textbf{Method}}
& \multicolumn{3}{c}{\textbf{Tiny-ImageNet}}    \\ 
\cline{3-5}
&             & T=2  & T=5   & T=10               \\ 
\hline
\multirow{6}{*}{\begin{tabular}[c]{@{}c@{}}$AvF. \downarrow$\end{tabular}}
& GM          &  5.05   &   6.10      &  4.60   \\
& SimGCD$^\ast$       &  3.01   &    7.53     &  8.87   \\ 
\cline{2-5}
& AutoNovel   & 32.10    &   52.61      &    57.71      \\
& FRoST$^\dagger$    & 5.05 &   -  &-  \\
& FRoST       &  2.15   &   4.69      &     6.32     \\ 
\cline{2-5}
& Our         &  \textbf{0.83}   &  \textbf{2.32}       &  \textbf{2.26}     \\
\cline{1-5}
\multirow{6}{*}{\begin{tabular}[c]{@{}c@{}}$AvD. \uparrow$\end{tabular}}
& GM          &  36.70   & 38.54        & 36.86    \\
& SimGCD$^\ast$     &  34.80   &      53.22   &   51.80  \\ 
\cline{2-5}
& AutoNovel   &  24.10   &    9.92     &    5.62      \\
& FRoST$^\dagger$    & 33.00 &   -  &-  \\
& FRoST       &   49.80  &    48.52     &   44.14       \\ 
\cline{2-5}
&   Our         &    \textbf{50.20} &   \textbf{53.32}      &  \textbf{57.06}      \\
\hline
\end{tabular}
}

\label{tab:for_img}
\end{table}

\paragraph{\bf Implementation Details.}
Following~\cite{vaze2022generalized,wen2023parametric}, we adopt the ViT-B/16~\cite{dosovitskiy2020image} pre-trained with DINO~\cite{caron2021emerging} as the feature extractor and only fine-tune the last transformer block. 
Our FEA model is trained using the SGD optimizer with a mini-batch size of 128 and the learning rate decays with a cosine schedule.
At the base session, we train our model for 10 epochs with an initial learning rate of 0.1. During incremental discovery sessions, we train our model for 200 epochs. 
The initial learning rate for CIFAR-10 and CIFAR-100 is set to 0.1 and that for Tiny-ImageNet is set to 0.02. And we set the hyper-parameters $\lambda$, $\varepsilon$, $\omega$, and $e$ to 0.01, 0.1, 2, and 30, respectively. 
During training sessions, we utilize random cropping, random horizontal flipping, and color jittering as data augmentation. Besides, all image samples are processed with center cropping for testing.
All experiments are conducted using the NVIDIA GeForce RTX 3090 GPU.

\subsection{Comparison with the State-of-the-Art Methods}
Due to the new experimental protocols and evaluation metrics, the results from previous works cannot be directly compared. 
We reproduce several SOTA methods under the same experimental protocols, including the NCD method AutoNovel~\cite{han2021autonovel}, the CNCD methods FRoST~\cite{roy2022class}, GM~\cite{zhang2022grow}, and the GCD method simGCD~\cite{wen2023parametric}. 
For a fair comparison, we utilize the ViT-B/16 pre-trained with DINO as the feature extractor for all our baselines. 
As simGCD is not originally designed for continual learning, to adapt it for the CNCD setting, we introduce our proposed guide-to-novel term $D_{g}$ to train its classifier and store exemplars for known classes as in iCaRL~\cite{rebuffi2017icarl} to maintain performance on known classes. 
We refer to this combined method as SimGCD$^\ast$. 
It is noteworthy that we store 20 exemplars for each known class by using the herding algorithm~\cite{welling2009herding} for GM and SimGCD$^\ast$.
We conduct comparison experiments on CIFAR-10, CIFAR-100, and Tiny-ImageNet with several protocols, where the incremental classes are divided equally into T sessions.  
\paragraph{\bf Evaluation on CIFAR-10.} 
As shown in Table~\ref{tab:acc}, Table~\ref{tab:for_cifar}, and Figure~\ref{figs:acc} (a), our proposed FEA method exceeds the best baseline GM by 7.24\% and 4.25\% in terms of average accuracy for T = 1 and T = 2 protocols on CIFAR-10. 

\paragraph{\bf Evaluation on CIFAR-100.} 

As we can see from Table~\ref{tab:acc}, Table~\ref{tab:for_cifar}, and Figure~\ref{figs:acc} (b)\&(c), our FEA demonstrates obvious advantages over the state-of-the-art methods when both T=2 and T=5: (1) it achieves the highest average accuracies; (2) it obtains the highest average discovery accuracies; (3) it maintains the lowest average forgetting rates. 

\begin{figure*}[t]
	\centering

	\includegraphics[width=1\textwidth]{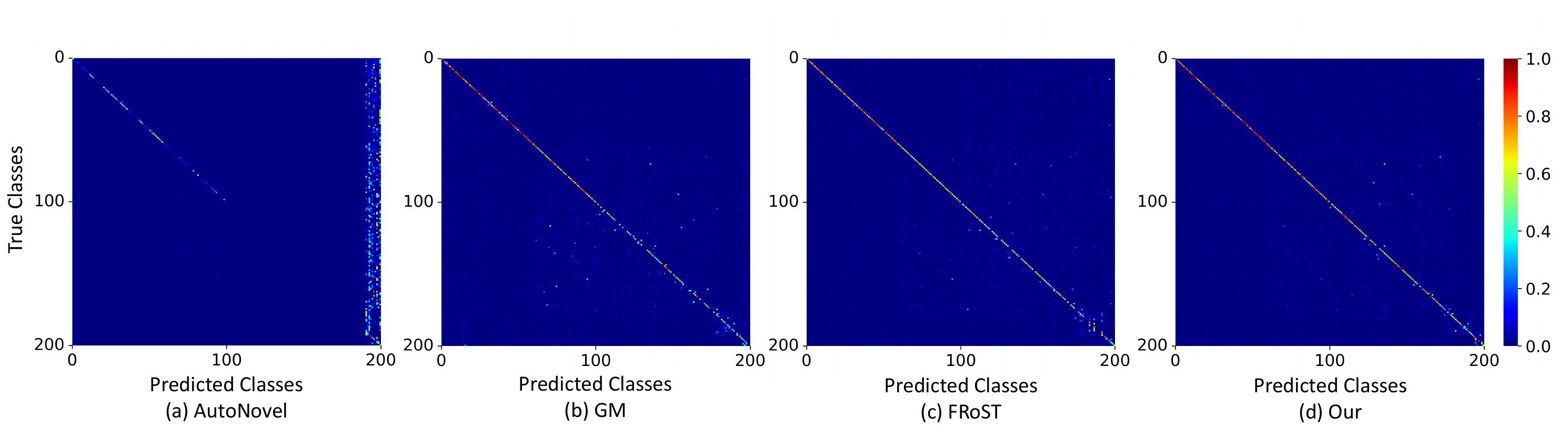} 
\caption{
The comparison of confusion matrices on Tiny-ImageNet under T=10 incremental protocol. The warmer color represents the higher accuracy.}
\label{figs:cm}
\end{figure*}

\paragraph{\bf Evaluation on Tiny-ImageNet.} To further verify the effectiveness of the FEA, we evaluate it on Tiny-ImageNet with more incremental sessions, which is more challenging for CNCD.
We can observe from Table~\ref{tab:acc}, Table~\ref{tab:for_img}, and Figure~\ref{figs:acc} (d)-(f), FEA outperforms the best existing method FRoST with a gap of 4.17\%, 5.11\%, and 7.92\% when T=2, T=5, and T=10 on average accuracy, respectively. Meanwhile, FEA surpasses the combined method SimGCD$^\ast$ by 2.95\%, 3.82\%, and 5.45\% when T=2, T=5, and T=10, respectively. 
Compared to FRoST and SimGCD$^\ast$, we find that with the increased number of incremental sessions, FEA exhibits a more pronounced advantage, showcasing stronger continual learning capability. 
Moreover, FEA obtains the lowest average forgetting rates and the highest average discovery accuracies compared to all methods.

\paragraph{\bf The comparison of the confusion matrices.} Figure~\ref{figs:cm} illustrates the comparison of confusion matrices achieved by AutoNovel, GM, FRoST, and our method FEA on Tiny-ImageNet under the T=10 protocol. 
The diagonal elements represent samples correctly classified, while off-diagonal elements represent samples misclassified. 
Due to the absence of measures to prevent catastrophic forgetting in AutoNovel, newly discovered classes and known classes are prone to be forgotten in subsequent sessions, and all samples tend to be classified as novel classes. 
The FEA achieves the best overall performance, not only maintaining the best recognition capability for known classes but also possessing the strongest ability to discover novel classes.

\subsection{Ablation Studies}

\begin{table}[]
\caption{Ablation study of each component in our method on CIFAR-10 and CIFAR-100.}
\renewcommand{\arraystretch}{1.28}
\centering
\resizebox{1.0\columnwidth}{!}{
\begin{tabular}{c|c|cc|cc|cc}
\hline

\multirow{2}{*}{\textbf{Metric}} & \multirow{2}{*}{\textbf{Method}} & \multicolumn{1}{c|}{\multirow{2}{*}{\textbf{CSS}}} & \multirow{2}{*}{\textbf{BAP}} & \multicolumn{2}{c|}{\textbf{Cifar-10}} & \multicolumn{2}{c}{\textbf{Cifar-100}} \\

\cline{5-8}    &        & \multicolumn{1}{c|}{}         &             & T=1           & T=2         & T=2         & T=5              \\ 
\hline
\multirow{4}{*}{\begin{tabular}[c]{@{}c@{}}$Avg.$ \\ $acc.\uparrow$\end{tabular}} 
&Framework      &   \ding{55}     &    \ding{55}           & 87.03              & 89.26             & 81.47              & 75.22             \\
& Framework + CSS      &   \ding{51}  &    \ding{55}           & 90.10              & 90.74             & 82.88              & 77.48             \\
& Framework + BAP       &   \ding{55}     &      \ding{51}       & 88.47              & 91.12             & 82.91              & 76.09             \\
& FEA       &   \ding{51}   &     \ding{51}      & \textbf{92.66}              & \textbf{92.30}             & \textbf{83.44}              & \textbf{79.09}             \\
\hline

\end{tabular}
}

\label{tab:ab}
\end{table}

\paragraph{\bf Component Analysis.}
The proposed FEA is comprised of three components:
a guide-to-novel framework, a centroid-to-samples similarity Constraint (CSS), and a boundary-aware prototype constraint (BAP). To prove the effectiveness of these components, we design ablation studies on CIFAR-10 and CIFAR-100. As we can see from Table~\ref{tab:ab}: (1) Our framework can handle the CNCD task independently and achieve satisfactory average accuracies, which demonstrates the superiority of our single-head paradigm. (2) The CSS and BAP can individually improve the performance of the framework, demonstrating the effectiveness of these two components. (3) The FEA achieves the highest accuracy, proving that three components can collaborate effectively to achieve optimal performance. 

\begin{table}[]
\caption{
        Comparison of the average accuracy with different settings of hyper-parameter
        $\omega$ on Cifar-100 under T=5 incremental protocol.}
\renewcommand{\arraystretch}{1.2}
\centering
\resizebox{0.9\columnwidth}{!}{
\begin{tabular}{c|ccccc}
\hline
$\omega$ & 0 &0.5& 1 & 2  & 4 \\ 
\hline
$Avg. acc. \uparrow$  & 77.49 &77.98 & 78.55  & \textbf{79.09}   & 77.87   \\ 
\hline
\end{tabular}
}

\label{tab:ab_w}
	
\end{table}

\paragraph{\bf Effectiveness of the boundary-aware prototype constraint.}
The hyper-parameter $\omega$ is used to control the weight of the BAP. We change the value of $\omega$ from 0 to 4. As indicated in Table~\ref{tab:ab_w}, our FEA method achieves the best performance when $\omega=2$. Moreover, compared to the performance of $\omega=0$, it proves the effectiveness of the BAP.

\section{CONCLUSIONS AND FUTURE WORKS}
In this paper, we propose a novel Feature Enhancement and Adaptation (FEA) method, which consists of a guide-to-novel framework following a single-head paradigm, a centroid-to-samples similarity constraint (CSS), and a boundary-aware prototype constraint (BAP). More specifically, to address the
feature-discrepancy problem, the guide-to-novel framework is established to continually discover novel classes under the guidance of prior distribution. Afterward, for feature enhancement, the CSS is designed to enhance the distinctiveness between novel class features. Finally, for feature adaptation, the BAP is proposed to better adapt novel class features to the shared feature space. We evaluate our method on CIFAR-10, CIFAR-100, and Tiny-ImageNet, in terms of performance comparisons, and ablation studies, demonstrating the effectiveness and superiority of our method.
A future research direction is to extend the CNCD method to cross-domain situations, that is, after supervised learning on one domain, continual novel
class discovery is carried out on other domains.

{\small
\bibliographystyle{ieee_fullname}
\bibliography{egbib}
}

\end{document}